\documentclass[journal]{IEEEtran}

\usepackage{epsfig} 
\usepackage[cmex10]{amsmath}
\usepackage{amssymb}  
\usepackage{bm} 
\usepackage{mathtools}
\usepackage[usenames,dvipsnames,svgnames]{xcolor}
\usepackage{booktabs} 
\usepackage{cite}
\usepackage{hyperref}
\makeatletter
\def\BState{\State\hskip-\ALG@thistlm}
\makeatother

\newtheorem{lemma}{Lemma}

\newtheorem{proposition}{Proposition}

\newtheorem{assumption}{Assumption}

\newcommand{\dif}{\mathrm{d}}

\newcommand{\sat}{\mathrm{sat}}

\begin{document}
%
\title{Safety-Augmented Operation of Mobile Robots Using Variable Structure Control}
%
%
%

\author{Azad Ghaffari and Seyed Amir Hosseini Dastja
\thanks{Azad Ghaffari is with the Department of Mechanical Engineering, Wayne State University, MI 48202, USA, {\tt\small aghaffari@wayne.edu}. Seyed Amir Hosseini Dastja is with EPSI and Danieli Automation, Isfahan, Iran, {\tt\small s.amir.hossaini@gmail.com}.} }

\maketitle


\begin{abstract}
The design process and complexity of existing safety controls are heavily determined by the geometrical properties of the environment, which affects the proof of convergence, design scalability, performance robustness, and numerical efficiency of the control. Hence, this paper proposes a variable structure control to isolate the environment's geometrical complexity from the control structure. A super-twisting algorithm is used to achieve accurate trajectory tracking and robust safety control. The safety control is designed solely based on distance measurement. First, a nominal safety model for obstacle avoidance is derived, where realistic system constraints are considered. The nominal model is well-suited for safety control design for obstacle avoidance, geofencing, and border patrol with analytically proven stability results. The safety control utilizes distance measurement to maintain a safe distance by compensating the robot's angular velocity. A supervisory logic is constructed to guarantee the overall stability and safety of the system. Operational safety and precision tracking are proven under parametric uncertainty and environmental uncertainty. The proposed design is modular with minimal tuning parameters, which reduces the computational burden and improves the control scalability. The effectiveness of the proposed method is verified against various case studies.
\end{abstract}

\section{Introduction}

Obstacle avoidance is an integral part of safety control for nonholonomic mobile robots, self-driving cars, unmanned aerial vehicles, and surface vehicles \cite{Peng2020, Mancini2019, Taghavifar2019, Liang2019, ghommam2019adaptive, behjat2019learning}. The literature includes various techniques such as potential field~\cite{Pan2019, karkoub2019trajectory, huang2006visual, pathak2005integrated, mas2011obstacle, rodriguez2014trajectory}, collision cone~\cite{fiorini1998motion, chakravarthy1998obstacle, fox1997dynamic, qu2004new, alonso2018cooperative}, path planning~\cite{fareh2020investigating, chu2012local, lamiraux2004reactive, divelbiss1997path}, model predictive control~\cite{Falcone2007, li2019dynamic, Pinkovich2020, Zhang2020}, and sliding-based method~\cite{matveev2011method, matveev2013problem}. Among early works, one can refer to the papers by Khatib~\cite{khatib1986real}, Krogh~\cite{krogh1985guaranteed}, Aggarwal, Leitmann, Skowronski~\cite{leitmann1977avoidance, aggarwal1972avoidance}, and Borenstein and Koren~\cite{borenstein1989real}. Recent developments include the usage of barrier certificates and control barrier functions to design safety controllers~\cite{Kong2013, Romdlony2016, Ames2017, glotfelter2019hybrid, Ghaffari2018, Ghaffari2020}.

Control design for autonomous vehicles needs to satisfy at least two objectives, such as trajectory tracking and obstacle avoidance. Thus, methods based on artificial potential field, kinodynamic motion planning, model predictive control, and barrier functions have been used in safety control design. However, it is known that the number, distribution, and shape of obstacles and mobile robots affect the analytical and computational complexity of the control design. For example, geometrical properties and spatial distribution of the obstacles directly affect the control structure and dramatically complicate proof of global convergence for such algorithms.

Popular methods such as artificial potential field and barrier functions require careful modeling of each object in the operational environment as a field or barrier function. In the case of artificial potential fields, the designer must foresee trap situations. On the other hand, barrier certificates must satisfy Lyapunov-like conditions, which render the application of the method very limited. Moreover, methods based on kinodynamic motion planning do not lead to exact solutions and are computationally intensive. 

Sliding-based methods have been proven effective in the design of safety-augmented control of autonomous robots~\cite{matveev2011method, matveev2013problem, Mancini2019, rodriguez2014trajectory}. Therefore, this paper provides control-oriented models and standardized design instructions based on a super-twisting algorithm to obtain a variable structure control with guaranteed safety features. The proposed design isolates the properties of the environment from the control design. The proposed control has a predefined structure with minimal control parameters, which conveniently handles multiple safety control problems, including obstacle avoidance, geofencing, and border patrol. The super-twisting algorithm is chosen as the primary control module to achieve precision trajectory tracking and safe operation. The super-twisting algorithm is robust, guarantees finite-time convergence, has a minimal set of parameters with explicit stability bounds, and features smooth transient and perfect tracking behavior.

This work focuses on nonholonomic mobile robots, which are driven by a differential drive. It is assumed that the thrust and torque are limited, and the linear velocity is positive. Thus, the robot always moves forward. The angular velocity, on the other hand, can take positive or negative values. Feedback linearization and the super-twisting algorithm are used to achieve accurate, robust trajectory tracking. The reference trajectory is known a priori and may cross obstacles or other robots. Also, in geofencing applications where a virtual safety net is introduced, the reference trajectory may leave the geofence for some periods. Moreover, only distance measurements, from obstacles, robots, or geofence boundaries are available to each mobile robot.

A framework based on distance measurement is presented for control design with augmented safety. Two system models have been obtained, one for trajectory tracking and one for safety control. The model used to design the trajectory tracking control is obtained using feedback linearization, with maximized control actuation along the transformed axes. Then, the super-twisting algorithm is used to guarantee precision trajectory tracking. Convergence is proven analytically. The safety control is designed using the dynamic model of distance variation. Actual system constraints, including actuator saturation, bounded velocity, and environmental characteristics, are used to obtain a nominal model suitable for safety control. The super-twisting algorithm compensates the effect of system uncertainties and environmental disturbances and maintains a safe distance between the robot and stationary or moving obstacles. When the safety control is active, the linear reference speed is modified such that the robot shadows the lead point of the reference trajectory. For small obstacles, however, the linear speed can be set to a fixed value.

A supervisory algorithm is designed to schedule the switching logic between trajectory tracking and safety control. The switching logic uses distance measurements, current robot position, and reference trajectory data to determine the time instants to switch back and forth between the two control modules. The supervisory algorithm handles obstacle avoidance and geofencing applications. The obstacle avoidance covers stationary and moving obstacles. The safety control can be used for border patrol applications. 

The proposed modular safety control dramatically improves control scalability by utilizing the super-twisting algorithm. Regardless of the number and planar distribution of the obstacles or complexity of the reference trajectory, the proposed algorithm maintains the safe trajectory tracking for the nonholonomic mobile robot. The algorithm's analytical complexity and required processing power are linearly scaled when the number of obstacles increases in the experiment.

The rest of this paper is presented in the following order. Section~\ref{sec:pre} presents preliminary and system modeling. System linearization and allowable control bounds are obtained in Section~\ref{sec:lin}. Trajectory tracking is explained in Section~\ref{sec:trtr}. The safety control is presented in Section~\ref{sec:safety}. The supervisory algorithm is discussed in Section~\ref{sec:sup}. Section~\ref{sec:sim} presents numerical simulations to verify the effectiveness of the proposed method. Section~\ref{sec:con} concludes the paper.

\section{Preliminary and System Modeling}\label{sec:pre}

\begin{figure}
\begin{center}
\includegraphics[clip,width=0.6\columnwidth]{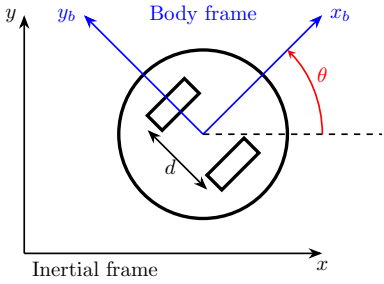}
\end{center}
\vspace{-5mm}
\caption{Nonholonomic mobile robot and its reference frames}
\label{fig:nmr}
\end{figure}
The nonholonomic mobile robot is driven by a differential drive comprised of two identical electric wheels. Fig.~\ref{fig:nmr} shows the schematic of the robot and inertial and body reference frames. The body reference frame is attached to the robot at the center of mass. The heading angle is $\theta$, which is measured with respect to the $x$-axis. The kinematic equations of the robot are given as
\setlength{\arraycolsep}{1pt}%
\begin{eqnarray}
\label{eq:dx}
\dot{x}&=&v\cos\theta\\
\label{eq:dy}
\dot{y}&=&v\sin\theta\\
\label{eq:dtheta}
\dot\theta&=&\omega,
\end{eqnarray}
where $p=[x~~y]^ T $ is the robot's position in the inertial reference frame, $ \theta $ is the heading angle, $v$ is linear velocity, and $\omega$ is angular velocity. The force and torque generated by the electric wheels control the linear and angular velocity of the robot. Assuming that the left and right wheel's movement does not affect each other, one can decouple the dynamic equation of the left and right wheels. Also, the effect of friction and wheel slip on the dynamic model is negligible. Since the mobile robot is symmetric with identical left and right wheels and motors, it is reasonable to assume that both wheels have the same dynamic equation as shown in the following
\begin{eqnarray}
\label{eq:vRL}
\dot{v}_R&=&-av_R+bu_R+\Delta_R\\
\dot{v}_L&=&-av_L+bu_L+\Delta_L,
\end{eqnarray}
where $a$ and $b$ are positive constants, and $v_R, v_L, u_R,$ and $u_L$ are the right wheel speed, left wheel speed, right motor voltage, and left motor voltage, respectively. Also, $ \Delta_R $ and $ \Delta_L $ are perturbation terms representing model uncertainty and external disturbance for the right and left wheels. The perturbation terms are bounded and Lipschitz continuous. The Robot's linear and angular velocity is related to the wheels' velocity through the following equations
\begin{eqnarray}
\label{eq:vu}
v&=&\big(v_R+v_L\big)/2\\
\label{eq:wu}
\omega&=&\big(v_R-v_L\big)/d,
\end{eqnarray}
where $d$ is the distance between the center of the wheels. Thus, the dynamic equations governing linear and angular velocity are obtained as
\begin{eqnarray}
\label{eq:dv}
\dot{v}&=&-av+bu_v+\Delta_v\\
\label{eq:domega}
\dot\omega&=&-a\omega+bu_\omega+\Delta_\omega,
\end{eqnarray}
where
\begin{eqnarray}
\label{eq:uv}
u_v&=&\big(u_R+u_L\big)/2\\
\label{eq:uomega}
u_\omega&=&\big(u_R-u_L\big)/d\\
\Delta_v&=&\big(\Delta_R+\Delta_L\big)/2\\
\Delta_\omega&=&\big(\Delta_R-\Delta_L\big)/d,
\end{eqnarray}
where $\Delta_v$ and $\Delta_\omega$ are bounded and Lipchitz continuous. Thus, the dynamic model of the mobile robot can be written as the following
\begin{eqnarray}
\label{sys:org}
\left[\begin{array}{c}\dot{x}\\\dot{y}\\\dot{\theta}\\\dot{v}\\\dot{\omega}\end{array}\right]=
\left[\begin{array}{c}v\cos\theta\\ v\sin\theta\\ \omega\\ -av+bu_v+\Delta_v\\ -a\omega+bu_\omega+\Delta_\omega\end{array}\right].
\end{eqnarray}
The voltage of the electric wheels are bounded as $|u_i|\le U$ for $i=R, L$, where $0<U< a^2d/b$. Using \eqref{eq:uv} and \eqref{eq:uomega}, one gets
\begin{eqnarray}
\label{eq:Uv}
|u_v|&\le& U\\
\label{eq:Uomega}
|u_\omega| &\le& {2U}/{d}.
\end{eqnarray}
Consider the nominal model, where $\Delta_v=\Delta_\omega=0$ in \eqref{sys:org}. Hence, if the initial condition satisfies $|v(0)|\le bU/a$ and $|\omega(0)|\le 2bU/(ad)$, then one can show that
\begin{eqnarray}
\label{eq:V}
|v(t)|&\le& bU/a\\
\label{eq:Omega}
|\omega(t)|&\le& 2bU/(ad),
\end{eqnarray}
for all $t>0$.

The model of mobile robot for safety control and trajectory tracking can be transformed into combination of first- and second-order linear differential equations with perturbations. Thus, one can use the super-twisting algorithm (STA) to design the trajectory tracking and safety control. The design steps of the STA is explained. Consider the following system
\setlength{\arraycolsep}{2pt}%
\begin{eqnarray}
\begin{array}{ccl}
\dot{z}_1&=&{z}_2\\
\dot{z}_2&=&w+\Delta(t),
\end{array}
\end{eqnarray}
where $\Delta(t)$ represents perturbations, including external disturbances and model uncertainties, which may be functions of time $t$. Assume $\Delta(t)$ is globally bounded and Lipschitz continuous, i.e., 
\begin{eqnarray}
\label{ineq:D1}
|\Delta(t)|&\le& M_\Delta\\
\label{ineq:D2}
\left| \dot{\Delta}(t)\right|&\le& L_\Delta,
\end{eqnarray}
where $M_\Delta$ and $L_\Delta$ are some known positive constants. The control objective is to track a reference value, $z_r$, which is enough smooth. Denote the error variables as $e_1=z_1-z_r$ and $e_2=z_2-\dot{z}_r$. Hence, the error dynamics are obtained as
\begin{eqnarray}
\begin{array}{ccl}
\dot{e}_1&=&e_2\\
\dot{e}_2&=&w-\ddot{z}_r+\Delta(t).
\end{array}
\end{eqnarray}

Consider a sliding surface as $\sigma=e_2+\lambda e_1$, where $\lambda>0$. The super-twisting algorithm is then designed as
\begin{equation}\label{eq:stc}
w=w^{\mathrm{eq}}-k_1\sqrt{|\sigma|}\mathrm{sign}(\sigma)-k_2\int_0^t\mathrm{sign}(\sigma)\dif\tau,
\end{equation}
where $k_2>L_\Delta$, $k_1>2\sqrt{k_2}$~\cite{Derafa2010, moreno2009linear}, and the equivalent control, $w^\mathrm{eq}$, is obtained from $\dot\sigma=0$ as
\begin{equation}
w^\mathrm{eq}=\ddot{z}_r-\lambda e_2.
\end{equation}
Moreover, if $|w|\le M_w$, one can use an integrator anti-windup to improve the transient behavior of the STA.

\section{Linearization and Allowable Control Bounds}\label{sec:lin}

The first control objective is to track a reference trajectory accurately. The super-twisting algorithm guarantees robust tracking performance with desirable transient and steady-state performance. First, the system dynamics are linearized using feedback linearization. Denote the following change of variables
\begin{eqnarray}
\label{eq:tr1}
\left[\eta_1~~\xi_1\right]^T&=&\left[x~~y\right]^T+R_\theta\left[L~~0\right]^T\\
\label{eq:tr3}
\left[\eta_2~~\xi_2\right]^T&=&R_\theta\left[v~~L\omega\right]^T,
\end{eqnarray}
where $L>0$ is a positive constant, and $R_\theta$ is the rotation matrix given as
\begin{equation}
\label{eq:Rtheta}
R_\theta=\left[\begin{array}{lr}\cos\theta~~&-\sin\theta\\\sin\theta&\cos\theta\end{array}\right].
\end{equation}
Also, two new control inputs $u_\eta$ and $u_\xi$ are defined as the following
\begin{eqnarray}
\label{eq:utr}
\left[\begin{array}{c}u_\eta\\ u_\xi\end{array}\right]=
R_\theta
\left[\begin{array}{c}-L\omega^2-av+bu_v\\v\omega-aL\omega+bLu_\omega\end{array}\right].
\end{eqnarray}
Applying \eqref{eq:tr1}--\eqref{eq:tr3}, with \eqref{eq:utr} as the new control inputs, the original dynamic equation \eqref{sys:org} is transformed into
\begin{eqnarray}
\label{sys:tr}
\left[\begin{array}{c}\dot\eta_1\\ \dot\eta_2\\ \dot\xi_1\\ \dot\xi_2\\ \dot\theta\end{array}\right]=
\left[\begin{array}{c}\eta_2\\ u_\eta+\Delta_\eta\\ \xi_2\\ u_\xi+\Delta_\xi \\ \left(\xi_2\cos\theta-\eta_2\sin\theta\right)/L \end{array}\right],
\end{eqnarray}
where
\begin{eqnarray}
\left[\begin{array}{c}\Delta_\eta\\\Delta_\xi\end{array}\right]=
R_\theta
\left[\begin{array}{c}\Delta_v\\ L\Delta_\omega\end{array}\right].
\end{eqnarray}
Note that the perturbation terms are bounded and Lipschitz continuous. The transformed system is comprised of two double-integrators, and the angle appears as a zero dynamic.
To convert \eqref{sys:tr} to the original dynamic equations \eqref{sys:org}, one can use the following inverse transformation
\begin{eqnarray}
\left[x~~y\right]^T&=&\left[\eta_1~~\xi_1\right]^T-R_\theta\left[L~~0\right]^T\\
\left[v~~L\omega\right]^T&=&R_\theta^{-1}\left[\eta_2~~\xi_2\right]^T\\
\label{eq:uvuomega}
\left[\begin{array}{c}u_v\\Lu_\omega\end{array}\right]&=&\frac{1}{b}\left[\begin{array}{c}av+L\omega^2\\aL\omega-v\omega\end{array}\right]+\frac{1}{b}R_\theta^{-1}\left[\begin{array}{c}u_\eta\\u_\xi\end{array}\right].
\end{eqnarray}

Moreover, one can find the allowable range of the transformed control inputs for the nominal model, where $\Delta_\eta=\Delta_\xi=0$. Denote $[w_1~~w_2]^T=R_\theta^{-1}[u_\eta~~u_\xi]^T$. Thus, one can use \eqref{eq:dv} and \eqref{eq:domega} to transform \eqref{eq:uvuomega} to the following 
\begin{equation}
\label{eq:dvdomega}
\left[\begin{array}{c}\dot{v}\\L\dot\omega\end{array}\right]=\left[\begin{array}{c}L\omega^2\\-v\omega\end{array}\right]+\left[\begin{array}{c}w_1\\w_2\end{array}\right].
\end{equation}
Consider $|w_1|\le M_1$ and $|w_2|\le M_2$, where $M_1$ and $M_2$ are positive constants. It is desirable to find $M_1$ and $M_2$ such that the acceleration bounds are satisfied. The bounds of the right-hand side of \eqref{eq:dvdomega} are obtained as
\begin{eqnarray}
-M_1\le &w_1+L\omega^2& \le M_1+\frac{4Lb^2}{a^2d^2}U^2\\
-M_2-\frac{2b^2}{a^2d}U^2\le &w_2-v\omega&  \le M_2+\frac{2b^2}{a^2d}U^2.
\end{eqnarray}
Also, the allowable bounds of the left-hand side of \eqref{eq:dvdomega} are obtained as $|\dot{v}|\le 2bU$ and $|L\dot\omega|\le 4LbU/d$. Thus, the following must be satisfied
\begin{eqnarray}
M_1&\le& 2bU-\frac{4Lb^2}{a^2d^2}U^2\\
M_2&\le& \frac{4Lb}{d}U-\frac{2b^2}{a^2d}U^2.
\end{eqnarray}
To guarantee existence of positive $M_1$ and $M_2$, the following must hold
\begin{equation}
\label{eq:L}
\frac{bU}{2a^2}< L < \frac{a^2d^2}{2bU}.
\end{equation}
Note that since $U< a^2d/b$, there is always a value for $L$ that satisfies \eqref{eq:L}. Moreover, note that $\|R_\theta^{-1}\|_\infty\le 1$. Thus, the bounds on $u_v$ and $u_\omega$ are satisfied if the following holds
\begin{equation}
\label{eq:UL}
\Big\|[u_\eta~~u_\xi]^T\Big\|_\infty \le U'(L),
\end{equation}
where
\begin{equation}
U'(L)=\min\left\{2bU-\frac{4Lb^2}{a^2d^2}U^2, \frac{4Lb}{d}U-\frac{2b^2}{a^2d}U^2\right\}.
\end{equation}
Note that $L=d/2$ satisfies \eqref{eq:L} and gives the maximum bounds of the transformed control inputs as 
\begin{equation}
\label{eq:Umax}
U'(d/2)= 2bU\left(1-\frac{bU}{a^2d}\right).
\end{equation}
The following lemma summarizes the discussion on the control bounds.
\begin{lemma}\label{Lem:bounds}
Consider the system \eqref{sys:org}, where the following hold $|u_v|\le U$,  $|u_\omega| \le {2U}/{d}$, $|v(0)|\le bU/a$, $\omega(0)\le 2bU/(ad)$, where $0<U<a^2d/b$. Then, $|v(t)|\le bU/a$ and $|\omega(t)|\le 2bU/(ad)$ for all $t>0$ for the nominal system. Moreover, if \eqref{eq:L} and \eqref{eq:UL} hold for the transformed inputs, the acceleration bounds of the original system \eqref{sys:org} are satisfied. Also, $L=d/2$ maximizes the bounds of $u_\eta$ and $u_\xi$, where the maximum bound is given by  \eqref{eq:Umax}.
\end{lemma}

\section{Trajectory Tracking}\label{sec:trtr}
The following explains the steps to design a trajectory control using the super-twisting algorithm. The reference trajectory is constructed such that the nonholonomic condition is satisfied, i.e.,
\begin{eqnarray}
\label{eq:dxr}
\dot{x}_r&=&v_r\cos\theta_r\\
\label{eq:dyr}
\dot{y}_r&=&v_r\sin\theta_r\\
\label{eq:dthetar}
\dot\theta_r&=&\omega_r,
\end{eqnarray}
where the proper selection of $v_r$ and $\omega_r$, within the allowed bounds, creates a variety of reference trajectories. First, the reference values for the transformed dynamics are calculated using \eqref{eq:dxr}--\eqref{eq:dthetar} and \eqref{eq:tr1}--\eqref{eq:tr3}
\begin{eqnarray}
\label{eq:qr1}
\left[\eta_{r1}~~\xi_{r1}\right]^T&=&\left[x_r~~y_r\right]^T+R_{\theta_r}\left[L~~0\right]^T\\
\label{eq:qr2}
\left[\eta_{r2}~~\xi_{r2}\right]^T&=&R_{\theta_r}\left[v_r~~L\omega_r\right]^T.
\end{eqnarray}
Denote the error variables as $\tilde\eta_i=\eta_i-\eta_{ri}$ and $\tilde\xi_i=\xi_i-\xi_{ri}$ for $i=1,2$, and $\tilde{\theta}=\theta-\theta_r$. Thus, the error dynamics are obtained as
\begin{eqnarray}
\label{eq:err1}
\left[\begin{array}{c}\dot{\tilde\eta}_1\\ \dot{\tilde\eta}_2\\ \dot{\tilde\xi}_1\\ \dot{\tilde\xi}_2\\ \dot{\tilde{\theta}}\end{array}\right]=
\left[\begin{array}{c} \tilde\eta_2\\ u_\eta-\dot\eta_{r2}+\Delta_\eta\\ \tilde\xi_2\\ u_\xi-\dot\xi_{r2}+\Delta_\xi\\ \Psi\end{array}\right],
\end{eqnarray}
where
\begin{eqnarray}
\Psi\!=\!\frac{(\xi_{r2}+\tilde\xi_2)\cos(\theta_r\!+\!\tilde\theta)\!-\!(\eta_{r2}+\tilde\eta_2)\sin(\theta_r\!+\!\tilde\theta)}{L}\!-\!\omega_r,
\end{eqnarray}
where the bound of the control inputs is given by \eqref{eq:UL}. In the subsequent analysis $L=d/2$.

Using the presented method in Section~\ref{sec:pre}, one can design the super-twisting algorithm for the error dynamics \eqref{eq:err1}. The sliding surfaces are designed as
\begin{equation}
\label{eq:sq}
s_q=\tilde{q}_2+c_q\tilde{q}_1, \quad c_q>0, \quad q=\eta, \xi.
\end{equation}
The super-twisting algorithm guarantees that $s_q$ goes to zero in finite time, which means the following hold
\begin{eqnarray}
\label{eq:etaeq}
\left[\tilde{\eta}_1~~\tilde{\eta}_2\right]^T&=&\left[0~~0\right]^T\\
\label{eq:xieq}
\left[\tilde{\xi}_1~~\tilde{\xi}_2\right]^T&=&\left[0~~0\right]^T.
\end{eqnarray}
Therefore, at the equilibrium point, the zero dynamic equation transforms to
\begin{eqnarray}
\label{eq:Dtheta}
\dot{\tilde{\theta}}&=&\frac{\xi_{r2}\cos(\theta_r+\tilde\theta)-\eta_{r2}\sin(\theta_r+\tilde\theta)}{L}-\omega_r.
\end{eqnarray}
Using \eqref{eq:qr2}, one can expand and simplify \eqref{eq:Dtheta} to arrive at the following
\begin{equation}
\label{eq:Dtheta2}
\dot{\tilde\theta}=\frac{L\omega_r\cos\tilde\theta-v_r\sin\tilde\theta}{L}-\omega_r.
\end{equation}
Denote $\phi=\arctan\left(v_r/(L\omega_r)\right)$. Thus, \eqref{eq:Dtheta2} is further simplified as
\begin{equation}
\dot{\tilde{\theta}}=\frac{\sqrt{L^2\omega_r^2+v_r^2}}{L}\left(\cos\left(\tilde\theta+\phi\right)-\cos\phi\right)
\end{equation}
which has two equilibrium points at $\tilde\theta_{\mathrm{eq}}=0, -2\phi$. Applying the Jacobian linearization, one gets
\begin{eqnarray}
\dot{\tilde\theta}=\left\{\begin{array}{rcl}-{v_r}\tilde\theta/L,&\quad \text{if}\quad & \tilde\theta_{\mathrm{eq}}=0\\
{v_r}\tilde\theta/L, &\quad \text{if}\quad & \tilde\theta_{\mathrm{eq}}=-2\phi
\end{array}\right. .
\end{eqnarray}
Also, note that $v_r$ is a positive real value. Hence, the equilibrium $\tilde\theta_\mathrm{eq}=-2\phi$ is unstable. On the other hand, the equilibrium $\tilde\theta_\mathrm{eq}=0$ is asymptotically stable which means that the actual heading angle converges to the reference heading angle. Moreover, it immediately follows that $R_\theta=R_{\theta_r}$ since $\theta=\theta_r$. Hence, one can use \eqref{eq:etaeq}--\eqref{eq:xieq}, \eqref{eq:tr1}--\eqref{eq:tr3}, and \eqref{eq:qr1}--\eqref{eq:qr2} to prove that $[x~~y]^T=[x_r~~y_r]^T$ and $[v~~\omega]^T=[v_r~~\omega_r]^T$ at the equilibrium. Thus, the proposed controller features perfect trajectory tracking.

The following proposition summarizes the control design and stability results.
\begin{proposition}
Consider that Lemma~\ref{Lem:bounds} holds for the mobile robot system given by \eqref{sys:org}.  The reference trajectory is given by \eqref{eq:dxr}--\eqref{eq:dthetar}, where $v_r>0$. Consider the nonlinear transformation given by \eqref{eq:qr1} and \eqref{eq:qr2}, where $L=d/2$. The sliding surfaces are given by \eqref{eq:sq}, where $c_\eta=c_\xi>0$. Then, the STA asymptotically stabilizes the origin of \eqref{eq:err1} and the robot accurately tracks the reference trajectory.
\end{proposition}

In the next section, the problem of safety control is formulated using only distance measurement. A nominal safety model is obtained, and then the super-twisting algorithm is used to design the safety control.

\section{Safety Control}\label{sec:safety}
Recall the dynamic equation of the mobile robot given by \eqref{sys:org}. As the robot tracks a reference trajectory in an uncontrolled environment, stationary and moving objects may appear on the robot's path. Also, in geofencing applications, it is desirable to maintain the robot inside a virtual net even if the reference trajectory leaves the geofence. In border patrol applications, the robot is required to maintain a certain distance from the boundary of an arbitrary set. In all of these applications, the robot must maintain a safe distance with an object or reroute its path to maintain a safe distance with the set boundary. Thus, in the subsequent analysis, a nominal model is obtained to facilitate safety control based on distance measurements. Then, the super-twisting algorithm is used to design the safety control.

\begin{figure}
\begin{center}
\includegraphics[clip, width=0.85\columnwidth]{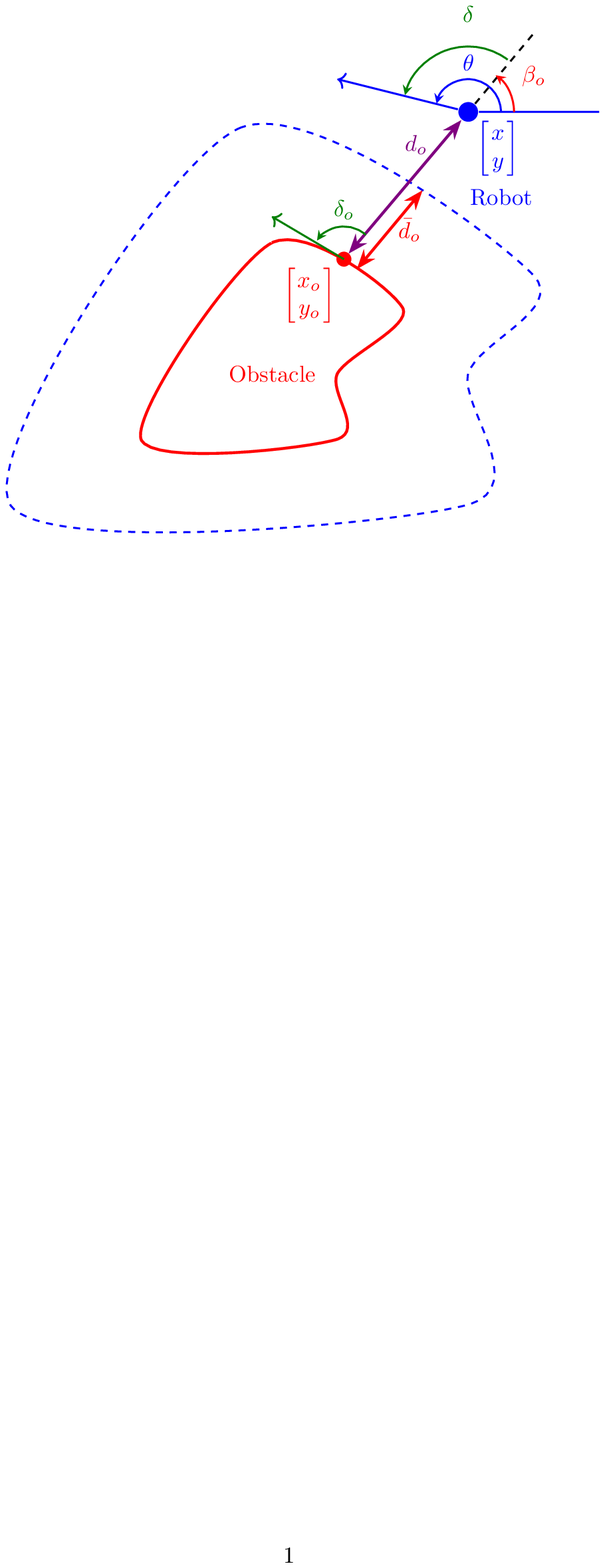}
\end{center}
\caption{An obstacle avoidance scenario is considered, where $d_o$ is the closest distance from the robot to the obstacle. It is desirable to keep the robot's distance at $\bar{d}_o$ from the obstacle.}
\label{fig:angles}
\end{figure}
Consider a mobile robot given by \eqref{sys:org}, where the robot's position is $p=[x~~y]^T$. Also, as shown in Fig.~\ref{fig:angles}, the closest point of the obstacle to the mobile robot is shown by $p_o=[x_o~~y_o]$. Since the location of $p_o$ changes when the robot moves, one can calculate the translational and rotational velocity of $p_o$ using the following relationships
\begin{eqnarray}
\label{eq:vo}
v_o&=&\sqrt{\dot{x}_o^2+\dot{y}_o^2}\\
\label{eq:omegao}
\omega_o&=&\frac{\ddot{y}_o\dot{x}_o-\ddot{x}_o\dot{y}_o}{\dot{x}_o^2+\dot{y}_o^2}.
\end{eqnarray}
As shown later, the knowledge of $v_o$ and $\omega_o$ is not required for the control design. However, the upper bound of $v_o$ and $\omega_o$ is required to prove the stability of the safety control. Also, since $v_o$ and $\omega_o$ are constructed using \eqref{eq:vo} and \eqref{eq:omegao}, then the following holds for $p_o$
\begin{eqnarray}
\label{eq:dxo}
\dot{x}_o&=&v_o\cos\theta_o\\
\label{eq:dyo}
\dot{y}_o&=&v_o\sin\theta_o\\
\label{eq:domegao}
\dot\theta_o&=&\omega_o.
\end{eqnarray}

Denote the length of $p_op$ as $d_o$ and its angle with the horizontal axis as $\beta_o$ calculated as
\begin{eqnarray}
\label{eq:do}
d_o&=&\sqrt{\left(x-x_o\right)^2+\left(y-y_o\right)^2}\\
\beta_o&=&\arctan\left(\frac{y-y_o}{x-x_o}\right).
\end{eqnarray}
Denote $\delta=\theta-\beta_o$ and $\delta_o=\theta_o-\beta_o$. Note that $\theta-\theta_o=\delta-\delta_o$. Next, take the derivative of $d_o$  and $\beta_o$ with respect to time.
\begin{eqnarray}
\dot{d}_o&=&\left(v\cos\theta-v_o\cos\theta_o\right)\frac{x-x_o}{d_o}+{}\nonumber\\
{}&&+\left(v\sin\theta-v\sin\theta_o\right)\frac{y-y_o}{d_o}\nonumber\\
{}&=&\left(v\cos\theta-v_o\cos\theta_o\right)\cos\beta_o+{}\nonumber\\
{}&&+\left(v\sin\theta-v\sin\theta_o\right)\sin\beta_o\nonumber\\
\label{eq:dotdo}
{}&=&v\cos\delta-v_o\cos\delta_o\\
\dot\beta_o&=&\left(v\sin\theta-v_o\sin\theta_o\right)\frac{\cos\beta_o}{d_o}-{}\nonumber\\
{}&&{}-\left(v\cos\theta-v_o\cos\theta_o\right)\frac{\sin\beta_o}{d_o}\nonumber\\
\label{eq:dbetao}
&=&\frac{v}{d_o}\sin\delta-\frac{v_o}{d_o}\sin\delta_o.
\end{eqnarray}
Taking the time derivative of $\dot{d}_o$ gives
\begin{eqnarray}
\label{eq:d2do}
\ddot{d}_o&=&\dot{v}\cos\delta-v\dot\delta\sin\delta-\dot{v}_o\cos\delta_o+v_o\dot\delta_o\sin\delta_o.
\end{eqnarray}
Note that $\dot\delta=\omega-\dot\beta_o$ and $\dot\delta_o=\omega_o-\dot\beta_o$, where $\dot\beta_o$ is given by \eqref{eq:dbetao}. Also, $\dot{v}$ is given by \eqref{sys:org}. Thus, one can write \eqref{eq:d2do} as the following
\begin{eqnarray}
\hspace{-5mm}
\ddot{d}_o&=&\big(-av+bu_v+\Delta_v\big)\cos\delta-v\omega\sin\delta+v_o\omega_o\sin\delta_o+\nonumber\\
{}&&{}+\dot\beta_o\big(v\sin\delta-v_o\sin\delta_o\big)-\dot{v}_o\cos\delta_o\nonumber\\
\label{eq:d2do2}
{}&=&\big(-av+bu_v\big)\cos\delta-v\omega\sin\delta+\frac{v^2}{d_o}\sin^2\delta+\Delta_o,
\end{eqnarray}
where
\begin{eqnarray}
\Delta_o&=&\Delta_v\cos\delta+v_o\omega_o\sin\delta_o+\frac{v_o^2}{d_o}\sin^2\delta_o-\nonumber\\
{}&&{}-\frac{2vv_o}{d_o}\sin\delta\sin\delta_o-\dot{v}_o\cos\delta_o.
\end{eqnarray}
Note that the robot linear and angular velocity are bounded. Also, assuming that the safe boundary around the obstacle is smooth enough such that \eqref{eq:vo} and \eqref{eq:omegao} is bounded and smooth enough, then one can find positive values $L_{o1}$ and $L_{o2}$ such that the following is satisfied
\begin{equation}
\label{eq:Do}
|{\Delta}_o|\le L_{o1}, \qquad |\dot\Delta_o|\le L_{o2}.
\end{equation}
The safety control is active when the robot's distance to the obstacle is within a determined safety bound, i.e., 
\begin{equation}
\label{eq:bardo}
\bar{d}_o-\epsilon_o\le d_o\le \bar{d}_o+\epsilon_o,
\end{equation}
where $0<\epsilon_o\ll\bar{d}_o$. Also, the angle $\delta$ is either $\pi/2$ for counterclockwise turns or $-\pi/2$ for clockwise turns. In the subsequent analysis, the counterclockwise scenario is addressed, i.e., 
\begin{equation}
\label{eq:bardelta}
\pi/2-\epsilon_\delta\le \delta \le \pi/2+\epsilon_\delta, 
\end{equation}
where $0<\epsilon_\delta\ll \pi/2$. Relationships between angles and distances are shown in Fig.~\ref{fig:angles}. Using \eqref{eq:bardo} and \eqref{eq:bardelta}, one can accurately approximate \eqref{eq:d2do2} as the following
\begin{eqnarray}
\ddot{d}_o&=&-v\omega+{v^2}/{\bar{d}_o}+\Delta'_o,
\end{eqnarray}
where $\Delta'_o=\Delta_o+\Delta_m$, where $\Delta_m$ accounts for the effect of model approximation, where $|\Delta'_o|<L_{o1}', |\dot\Delta'_o|<L_{o2}'$, where $L_{o1}'$ and $L_{o2}'$ are positive constants. Denote $\zeta_1=d_o-\bar{d}_o$ and $\zeta_2=\dot\zeta_1$, where $\bar{d}_o$ is the safe distance from the obstacle. Hence, the nominal safety model is given as
\begin{eqnarray}
\left[\begin{array}{c}\dot{\zeta}_1\\ \dot\zeta_2\\  \dot\omega \\  \dot{v} \\ \dot\delta \end{array}\right]=
\left[\begin{array}{c}\zeta_2 \\ -v\omega+v^2/\bar{d}_o+\Delta_o' \\ -a\omega+bu_\omega \\ -av+bu_v \\ \omega-\left(v\sin\delta-v_o\sin\delta_o\right)/\left(\bar{d}_o+\zeta_1\right)\end{array}\right].
\end{eqnarray}
During the safety control phase, the translational speed is set to the reference value, $v_r$, and the angular velocity is compensated to maintain the safe distance at $\bar{d}_o$. To design the safety control, first, the backstepping technique is used to drive $\zeta_1$ and $\zeta_2$ to zero.  Denote $\tilde\omega=\omega-\hat\omega_r$, $\tilde{v}=v-v_r$, and $\tilde\delta=\delta-\pi/2$, where
\begin{equation}
\label{eq:homegar}
\hat\omega_r={v}/{\bar{d}_o}-\hat{u}_\zeta/v.
\end{equation}
Also, define the control inputs as
\begin{eqnarray}
\label{eq:usomega}
u_\omega&=&\left(-{c_\omega}\tilde\omega+a\omega+\dot{\hat\omega}_r\right)/b\\
\label{eq:usv}
u_v&=&\left(-{c_v}\tilde{v}+av\right)/b,
\end{eqnarray}
where $c_v>0$ and $c_\omega>0$. Thus, the safety model is updated as
\begin{eqnarray}
\label{eq:errzeta}
\left[\begin{array}{c}\dot{\zeta}_1\\ \dot\zeta_2\\  \dot{\tilde\omega} \\  \dot{\tilde{v}} \\ \dot{\tilde\delta} \end{array}\right]=
\left[\begin{array}{c}\zeta_2 \\ \hat{u}_\zeta+\Delta_o'' \\ -c_\omega\tilde\omega \\ -c_v\tilde{v} \\ \Phi\end{array}\right],
\end{eqnarray}
where
\begin{equation}
\Phi=\hat{\omega}_r+\tilde\omega-\frac{(v_r+\tilde{v})\cos\tilde\delta-v_o\sin\delta_o}{\bar{d}_o+\zeta_1},
\end{equation}
where $\Delta_o''=\Delta_o'-v\tilde\omega$, where $|\Delta_o''|<L_{o1}''$ and $|\dot{\Delta}_o''|<L_{o2}''$, where $L_{o1}''$ and $L_{o2}''$ are positive constants. Note that $\hat\omega_r$ is an auxiliary input, which must satisfy \eqref{eq:Omega}. Thus, using \eqref{eq:homegar}, one arrives at
\begin{eqnarray}
|\hat\omega_r v|\le |v^2/\bar{d}_o|+|\hat{u}_\zeta|,
\end{eqnarray}
where
\begin{equation}
|\hat\omega_r v|-|v^2/\bar{d}_o|\le 2b^2U^2/(a^2d)-b^2U^2/(a^2\bar{d}_o).
\end{equation}
Therefore, the bounds on $\hat{u}_\zeta$ are obtained as
\begin{equation}
\label{eq:Uzeta}
|\hat{u}_\zeta|\le \left(bU/a\right)^2\left(2/d-1/\bar{d}_o\right),
\end{equation}
where $\bar{d}_o>d/2$, which verifies the fact that the safe distance, measured from the robot's center of mass, must be larger than the wheels' distance from the robot's center of mass. Otherwise, the robot will collide with the obstacle.

The super-twisting algorithm is used to design $\hat{u}_\zeta$. First, the sliding surface is designed as
\begin{equation}
\label{eq:szeta}
s_\zeta=\zeta_2+c_\zeta \zeta_1, \quad c_\zeta>0.
\end{equation}
Next, the super-twisting algorithm proves that $s_\zeta$ goes to zero in finite time which in turn proves that $\zeta_1$ and $\zeta_2$ converge to zero asymptotically. Hence, the robot is maintained at the safe distance, $\bar{d}_o$, from point $p_o$.

\begin{figure}
\begin{center}
\includegraphics[clip, width=0.9\columnwidth]{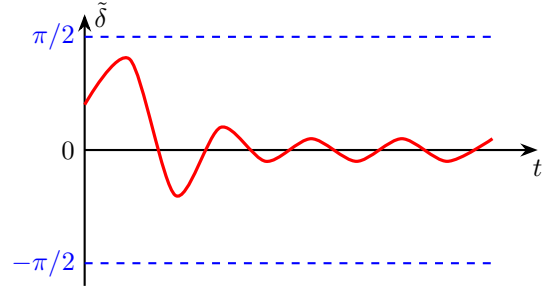}
\end{center}
\vspace{-5mm}
\caption{Variation of $\tilde\delta$ over time. The safety algorithm maintains $\tilde\delta\approx0$.}
\label{fig:tildedelta}
\end{figure}

Furthermore, the error system \eqref{eq:errzeta} shows that $\tilde\omega$ and $\tilde{v}$ converge to zero. Assume that the obstacle is a single stationary point, i.e., $v_o=0$. Thus, the zero dynamic can be written as
\begin{eqnarray}
\label{eq:dtildedelta}
\dot{\tilde\delta}&=&\hat{\omega}_r-\frac{v_r}{\bar{d}_o}\cos\tilde\delta.
\end{eqnarray}
The behavior of \eqref{eq:dtildedelta} depends on the sign of $\hat\omega_r$. If the robot is within the avoidance distance and $d_o>\bar{d}_o$, then $-\pi/2<\tilde\delta<0$. Thus, the super-twisting algorithm creates enough positive angular velocity to drive the distance toward $\bar{d}_o$. Therefore, $\tilde\delta\to0$. On the other hand, if $d_o<\bar{d}_o$, then $0<\tilde\delta<\pi/2$. Thus, the super-twisting algorithm creates enough negative angular velocity to drive the distance to $\bar{d}_o$, and thus $\tilde\delta \to 0$. The super-twisting algorithm continuously modifies the angular velocity to maintain the robot at the safe distance. Thus, as shown in Fig.~\ref{fig:tildedelta}, the two behaviors always are present, which means the zero dynamic is stable in an average sense, and thus $\tilde\delta\approx 0$. A similar argument can be used for the case when the obstacle is moving, or the problem of border patrol is addressed. Similar results also can be obtained for the case of clockwise safety control.

The following proposition summarizes the results of counterclockwise safety control.
\begin{proposition}
Consider the mobile robot modeled by \eqref{sys:org}, where $\Delta_v$ and $\Delta_\omega$ are bounded and Lipschitz continuous. Assume the robot is in the vicinity of the avoidance distance, i.e., $|d_o-\bar{d}_o|\le \epsilon_o$, where $0<\epsilon_o\ll d_o$. Moreover, the closest point of the obstacle to the robot changes as \eqref{eq:dxo}--\eqref{eq:domegao}, where $v_o$ and $\omega_o$ are bounded and enough smooth. The control law is given as \eqref{eq:homegar}--\eqref{eq:usv}, where $\hat{u}_\zeta$ is designed using the super-twisting algorithm. Then, the origin of the error dynamics \eqref{eq:errzeta} is stable and the robot maintains safe distance $\bar{d}_o$ with the obstacle.
\end{proposition}
\section{Supervisory Algorithm}\label{sec:sup}

The proposed control algorithm comprises trajectory tracking and safety control, where the stability of each module is separately proven. This section presents the supervisory algorithm that provides the switching logic between the two modules. The proposed supervisory algorithm can handle obstacle avoidance of stationary and moving objects, geofencing, and border patrol applications. 

The mobile robot dynamic equation is given by \eqref{sys:org}. Denote $X=[x~y~\theta~v~\omega]^T$, where $X\in\mathcal{X}$, where $\mathcal{X}\subseteq\mathbb{R}^5$. The reference trajectory is defined as $X_r=[x_r~y_r~\theta_r~v_r~\omega_r]^T$. Note that the reference trajectory satisfies the nonholonomic condition. A set of initial conditions $\mathcal{X}_0\subset\mathcal{X}$ and a set of unsafe states $\mathcal{X}_u\subset\mathcal{X}$ are given. The safety is achieved if all the state trajectories initiated inside $\mathcal{X}_0$ avoid the unsafe set for all $t>0$. The avoidance zone of an obstacle is a strip of a predefined width around the obstacle. In keep-in geofencing applications, the geofence's avoidance zone is a strip of predefined width around the inner border. The width of the avoidance zone may differ between obstacles. The safe set of the mobile robot is bound by the geofence borders and excludes the border's avoidance zone and all obstacles and their avoidance zones. 

It is desirable to design the supervisory algorithm such that the robot accurately tracks the reference trajectory inside the safe set and successfully avoids the unsafe set. Consider the following assumptions.
\begin{assumption}
The system starts from a safe state.
\end{assumption}
\begin{assumption} 
The stationary obstacles are far apart with non-intersecting avoidance zones. Also, the robot can navigate safely through spaces between neighboring obstacles.
\end{assumption}
\begin{assumption} 
At any given instant, no more than two robots are on a collision path.
\end{assumption}
\begin{assumption}
The obstacle dimensions are comparable to the robot dimensions.
\end{assumption} 

\begin{figure}
\begin{center}
\includegraphics[clip,width=0.85\columnwidth]{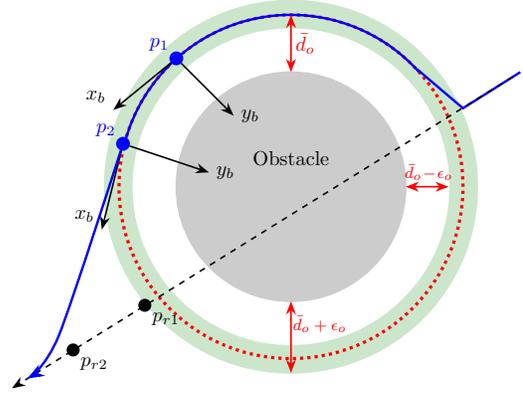}
\end{center}
\vspace{-3mm}
\caption{Obstacle avoidance scenario. The reference trajectory is shown in dashed black. The robot moves from north-east to south-west on the solid blue curve. If the trajectory control activated at $p_1$, excessive switching happens between trajectory tracking and safety control. The safe distance is $\bar{d}_o$ and $0<\epsilon_o\ll\bar{d}_o$. The green strip is the avoidance zone.}
\label{fig:PosEy}
\end{figure}

Consider counterclockwise safety control is implemented in the following discussion. Similar results can be produced for clockwise safety control. 
The position error in the robot body frame is obtained as
\begin{equation}
\left[e_x~~e_y\right]^T=R_\theta\left[x_r-x~~y_r-y\right]^T,
\end{equation}
where $R_\theta$ is the rotation matrix given by \eqref{eq:Rtheta}. Note that $e_y$ is the projection of position error vector along $y_b$-axis. See Fig.~\ref{fig:PosEy}.

Since the system initially is at a safe state, $\left\{X,X_r\right\}\subset\mathcal{X}_0$, the trajectory tracking algorithm is active when the experiment starts. If the reference trajectory and robot are in the vicinity of an obstacle or another robot, $\left\{X,X_r\right\}\subset\mathcal{X}_u$, the safety control is activated. The safety control may be deactivated when the reference trajectory leaves the avoidance zone. However, the robot's position relative to the reference trajectory affects the decision to reactivate the trajectory tracking. 

As Fig.~\ref{fig:PosEy} shows, the reference trajectory leaves the avoidance zone at point $p_{r1}$ while the robot is at point $p_1$. One can observe that the straight path to the reference trajectory intersects with the avoidance zone, i.e., $e_y>0$. Thus, if the safety control is deactivated, the robot may go back inside the avoidance zone. Hence, excessive switching between safety control and trajectory tracking may worsen system performance and harm the actuators. On the other hand, when the robot arrives at point $p_2$, where the corresponding reference point is at $p_{r2}$, one can observe that the straight line between the robot and reference point no longer crosses the avoidance zone. Thus, $e_y\le0$ indicates that if the safety control is deactivated, the robot leaves tangent to the avoidance boundary. Thus, the safety control can be turned off when the reference trajectory is outside the avoidance region and $e_y\le0$. Hence, excessive switching is avoided. If the clockwise safety control is implemented,  $e_y\ge0$ deactivates the safety control. 
\begin{figure}
\begin{center}
\includegraphics[clip, width=0.9\columnwidth]{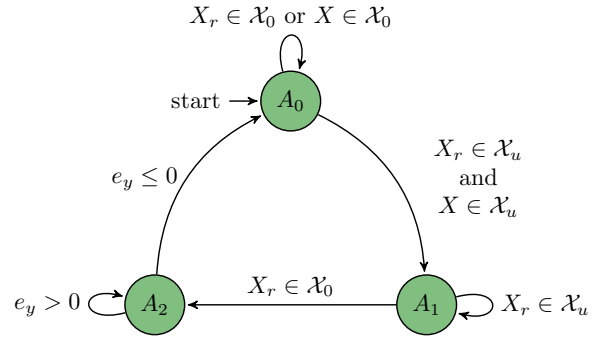}
\end{center}
\vspace{-3mm}
\caption{State machine for obstacle avoidance. The trajectory control is active in state $A_0$, and the safety control is active in states $A_1$ and $A_2$. If conditions $e_y>0$ and $e_y\le0$ are replaced with $e_y<0$ and $e_y\ge0$, respectively, the logic can be used for keep-in geofencing or border patrol.}
\label{fig:StateMachineA}
\end{figure}

The supervisory algorithm for obstacle avoidance is shown in Fig.~\ref{fig:StateMachineA}. Similar results can be produced for clockwise safety control. If the obstacle is oversize, the obstacle avoidance problem can be treated as a geofencing problem. In keep-in geofencing, it is desirable to maintain the robot inside the safe zone at a safe avoidance distance with the border regardless of the changes in the reference trajectory. The safe zone comprises a strip of width $\bar{d}_o$ around the inner border. If the conditions $e_y>0$ and $e_y\le0$ in Fig.~\ref{fig:StateMachineA} are replaced with $e_y>0$ and $e_y\ge0$, respectively, one obtains the supervisory algorithm with counterclockwise safety control for keep-in geofencing and border patrol applications.

\section{Control Implementation and Case Studies}\label{sec:sim}

Four case studies are designed to verify the effectiveness of the proposed safety control, including a) avoiding stationary obstacles, b) safe passage of colliding robots, c) keep-in geofencing, and d) border patrol. 

The trajectory tacking control is the same for the four case studies as given in the following
\begin{equation}
\label{eq:uq}
u_q=u_{eq}-k_{q1}\sqrt{|s_q|}\mathrm{sign}(s_q)-k_{q2}\int_0^t\mathrm{sign}(s_q)\dif\tau,
\end{equation}
where $s_q=\tilde{q}_2+c_q\tilde{q}_1$ for $q=\eta,\xi$. The equivalent control is given as
\begin{equation}
u_q^{eq}=\ddot{q}_r-c_q \tilde{q}_2, \quad q=\eta, \xi.
\end{equation}
The control parameters are set as $k_{q1}=2$, $k_{q2}=0.5$, $c_q=10$ for $q=\eta, \xi$.
Note that $L=d/2$. Thus, one can use \eqref{eq:UL} and \eqref{eq:Umax} to obtain the control bounds as $|u_q|\le U_q$, where 
\begin{equation}
U_q = 2bU\left(1-\frac{bU}{a^2d}\right), \quad q=\eta,\xi,
\end{equation}
where $a=b=3.85$, $d=0.235$~m, $U=0.7$~V.
The saturated control command is then calculated as $u_q^\mathrm{sat}=U_q\sat\left(u_q/U_q\right)$ for $q=\eta, \xi$, where the saturation function is defined as
\begin{equation}
\sat(\psi)=\left\{\begin{array}{lcl}\psi &\text{if}& |\psi|<1\\ \mathrm{sign}(\psi)&\text{if}& |\psi|\ge1\end{array}\right. .
\end{equation}
Since the control \eqref{eq:uq} includes an integrator and the control bounds are know, one can add an integrator anti-windup to improve the closed-loop performance of the system. Next, one can use \eqref{eq:uvuomega} to calculate $u_v$ and $u_\omega$ from $u_\eta^\mathrm{sat}$ and $u_\xi^\mathrm{sat}$, where $|u_v|\le U$, $u_\omega\le U_\omega$, where $U_\omega=2U/d$. Hence, the control command of the left and right wheels are obtained as
\begin{equation}
\left[\begin{array}{c}u_R\\u_L\end{array}\right]=\left[\begin{array}{lr}1&d/2\\ 1&-d/2\end{array}\right]\left[\begin{array}{c}U\sat\left(u_v/U\right)\\ U_\omega\sat\left(u_\omega/U_\omega\right)\end{array}\right].
\end{equation}
Because of the actuator limitation, the saturated command signals are applied to the wheels, i.e., $U\sat\left(u_R/U\right)$ and $U\sat\left(u_L/U\right)$.  

\begin{figure}
\begin{center}
{\bf (a)}\\
\includegraphics[width=0.7\columnwidth]{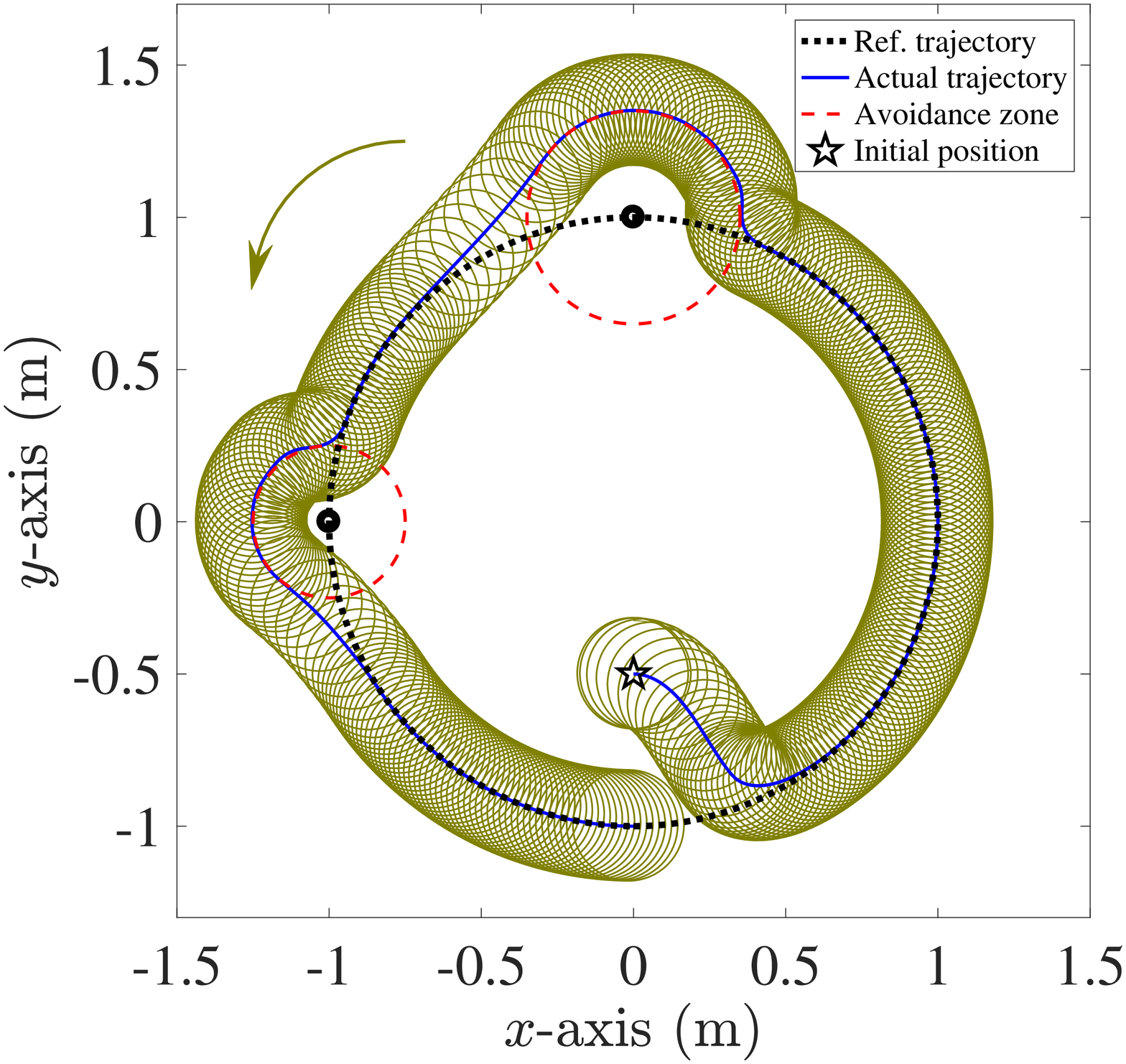}\\
{\bf (b)}\\
\includegraphics[width=0.9\columnwidth]{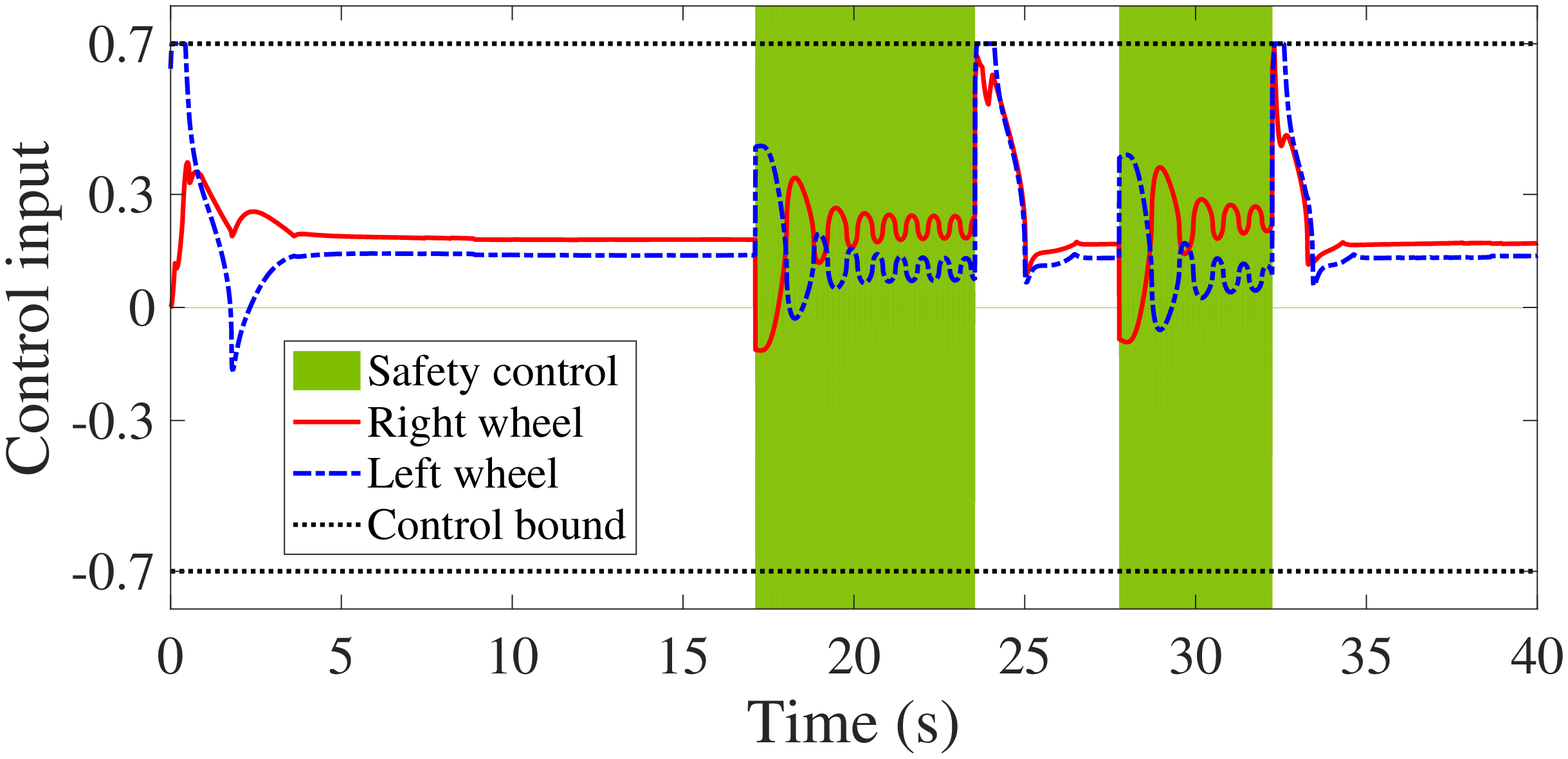}
\end{center}
\vspace{-3mm}
\caption{{\bf(a)} The robot accurately tracks the reference trajectory outside the avoidance zone and successfully avoids the stationary obstacles. Solid blue is the center of mass of the robot. {\bf (b)} Evolution of control commands versus time. The safety control is active in the green area.}
\label{fig:OApos}
\end{figure}

\subsection{Avoiding Stationary Obstacles}
Assume that the robot tracks a reference trajectory in an environment where stationary obstacles are present. The reference trajectory may cross the obstacles. The safety control is given as \eqref{eq:usomega}--\eqref{eq:usv}, where $\hat\omega_r$ is given by \eqref{eq:homegar}. However, as explained earlier in Fig.~\ref{fig:tildedelta}, $\tilde\delta$ does not settle at a fixed value, which means $\hat\omega_r$ features continuous change. Thus, to improve performance robustness, a modified version of $u_\omega$ is implemented as the following 
\begin{eqnarray}
u_\omega&=&\left(-{c_\omega}\tilde\omega+a\omega\right)/b,
\end{eqnarray}
where $\tilde\omega=\omega-\hat\omega_r$, where $\hat\omega_r$ is given by \eqref{eq:homegar}. The super-twisting algorithm is used to design $\hat{u}_\zeta$ as
\begin{equation}
\hat{u}_\zeta=-c_\zeta\zeta_2-k_{\zeta1}\sqrt{|s_\zeta|}\mathrm{sign}(s_\zeta)-k_{\zeta2}\int_0^t\mathrm{sign}(s_\zeta)\dif\tau,
\end{equation}
where $s_\zeta=\zeta_2+c_\zeta\zeta_1$. The control parameters are designed as $k_{\zeta1}=0.8$, $k_{\zeta2}=0.04$, $c_\zeta=1$, $c_u=c_\omega=5$. The bounds of $\hat{u}_\zeta$ are given by \eqref{eq:Uzeta}. Thus, an integrator anti-windup can be added to the safety algorithm to further improve the control performance. Also, to incorporate the control saturation, one can modify \eqref{eq:homegar} as 
\begin{equation}
\hat\omega_r={v}/{\bar{d}_o}-\sat\left(\hat{u}_\zeta/U_\zeta\right)/v,
\end{equation}
where $U_\zeta=\left(bU/a\right)^2\left(2/d-1/\bar{d}_o\right)$, where $\bar{d}_o$  is the safe distance.

Two stationary obstacles are considered at positions $p_1=[0~1]^T$ and $p_2=[-1~0]^T$, where the safe distance is $0.35$~m for $p_1$ and $0.25$~m for $p_2$. The robot is located at $p_0=[0~-0.5]^T$. As shown in Fig.~\ref{fig:OApos}(a), the robot accurately tracks the reference trajectory and successfully avoids the two obstacles. The green circles represent the robot's actual dimension, and the dashed red circles are the avoidance zones. Note that the robot is modeled as a point mass. The solid blue line represents the center of mass of the robot. The reference velocities are given as $v_r=0.5\pi$~m/s and $\omega_r=0.5\pi$~rad/s. The evolution of the control signals is shown in Fig.~\ref{fig:OApos}(b), where the green area shows the duration where the safety control is active. During the safety control, $v_r=0.5\pi$~m/s, and the angular velocity is manipulated to maintain the safe distance with the obstacle.

\subsection{Safe Passage of Colliding Robots}
\begin{figure}
\begin{center}
{\bf (a)}\\
\includegraphics[width=0.7\columnwidth]{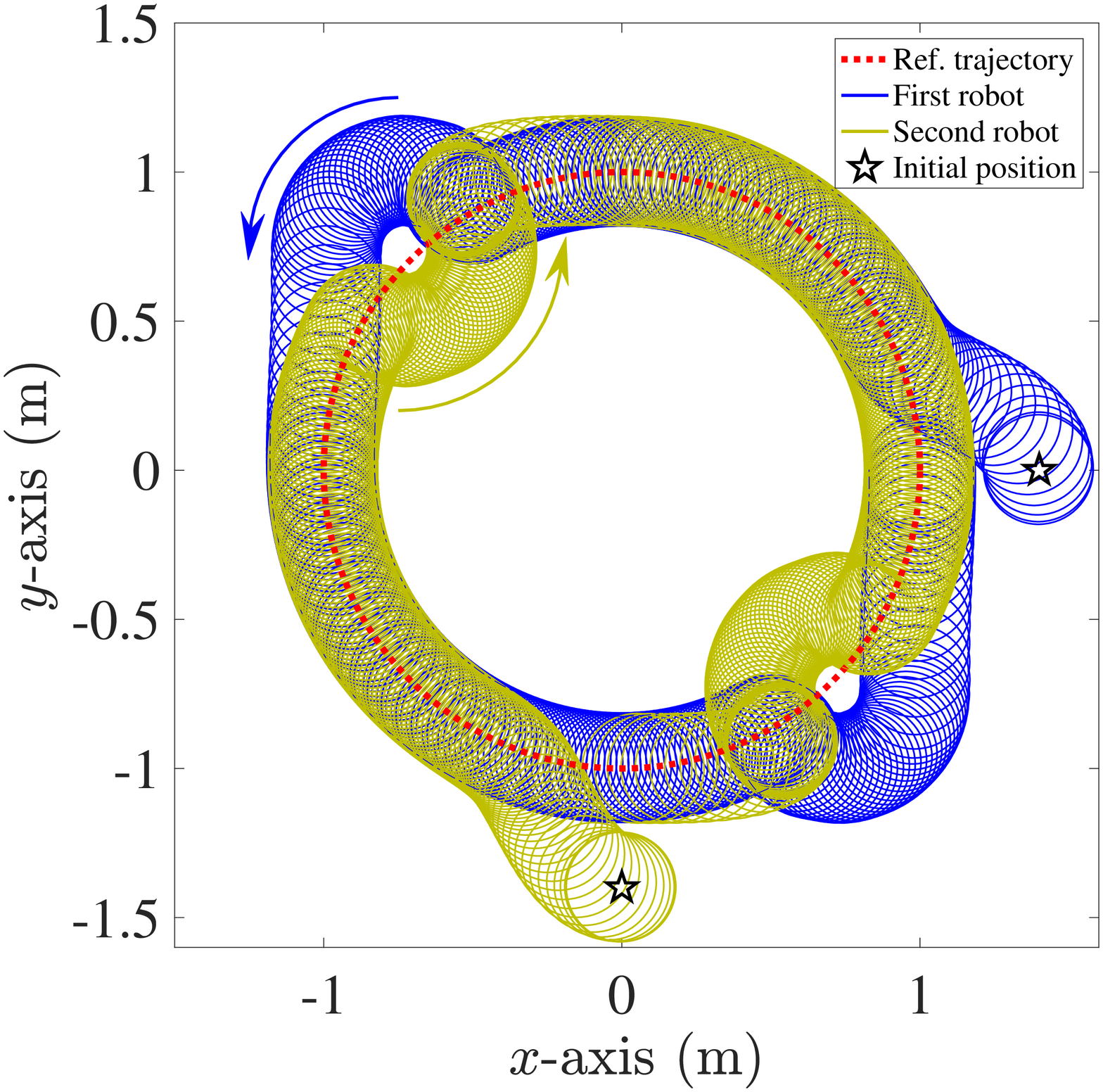}\\
{\bf (b)}\\
\includegraphics[width=0.9\columnwidth]{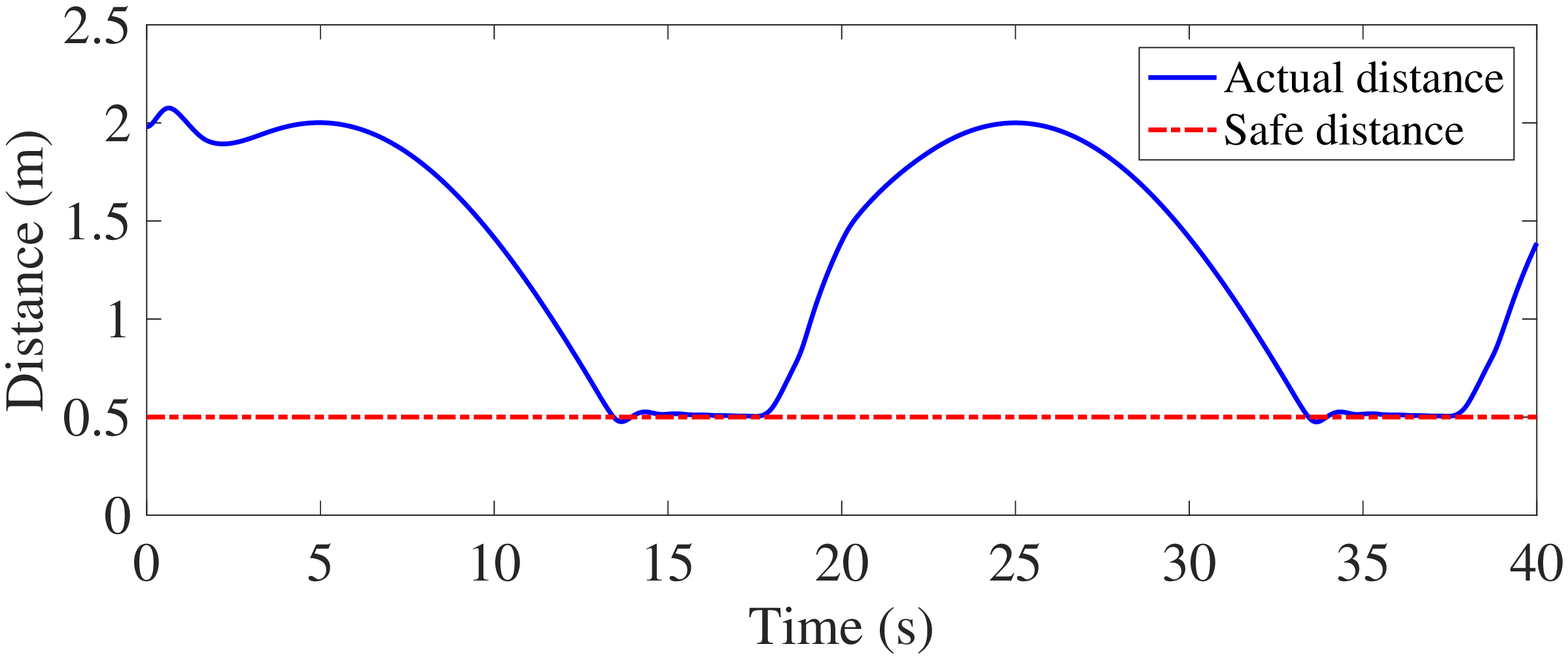}
\end{center}
\vspace{-3mm}
\caption{{\bf (a)} Two robots move on the same trajectory in opposite directions. The blue robot moves counterclockwise and the green robot moves clockwise. The proposed algorithm guarantees safe operation. {\bf (b)} Distance between the two robots. The robots do not violate the safety distance.}
\label{fig:CApos}
\end{figure}

Consider a situation where two robots are moving on the same trajectory in the opposite direction. Thus a collision is inevitable without safety control. Each robot is equipped with the same obstacle avoidance algorithm designed for avoiding stationary obstacles. As Fig.~\ref{fig:CApos}(a) shows, the robots safely pass each other. The robots are identical and $\bar{d}_o=0.5$~m for both robots. As shown in Fig.~\ref{fig:CApos}(b), the safe distance is kept at $\bar{d}_o$, although the reference trajectories result in a collision. The initial positions of the robots are $[1.4~~0]^T$~m and $[0~~-1.4]^T$~m.

\subsection{Keep-In Geofencing}
\begin{figure}
\begin{center}
\includegraphics[width=0.6\columnwidth]{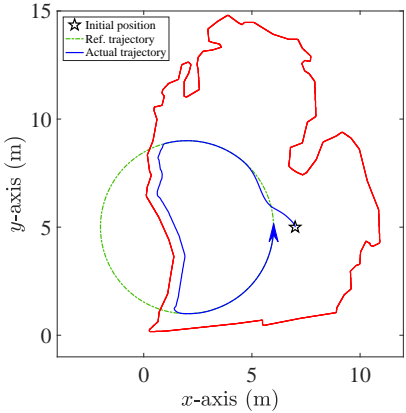}
\end{center}
\vspace{-3mm}
\caption{The proposed algorithm accurately tracks the reference trajectory while maintaining the robot at a safe distance inside the geo-fence shown in solid red.}
\label{fig:GFpos}
\end{figure}
The proposed algorithm can also handle the problem of geofencing and border patrol. The trajectory tracking and safety control have the same structure as the obstacle avoidance, except $c_\zeta=10$, $c_u=c_\omega=3$. Moreover, performance may be improved by modifying the linear reference speed during the safety phase such that the robot shadows the lead point of the reference trajectory. 

For example, for keep-in geofencing application, if the length of the reference trajectory outside the border, is longer than the safe border cropped by the reference trajectory, one can use the following
\begin{equation}
\hat{v}_r=v_r\cos\left(\theta-\theta_r\right),
\end{equation}
to modify the linear reference velocity. Providing a general modification algorithm for the linear reference velocity is outside the scope of this work. 

The geofencing scenario is shown in Fig.~\ref{fig:GFpos}. Although the reference trajectory leaves the border, the robot maintains a safe distance at $d_o=0.5$~m with the border. The linear speed is updated to keep up with the lead point of the reference trajectory. 

\subsection{Border Patrol}
\begin{figure}
\begin{center}
{\bf (a)}\\
\includegraphics[width=0.6\columnwidth]{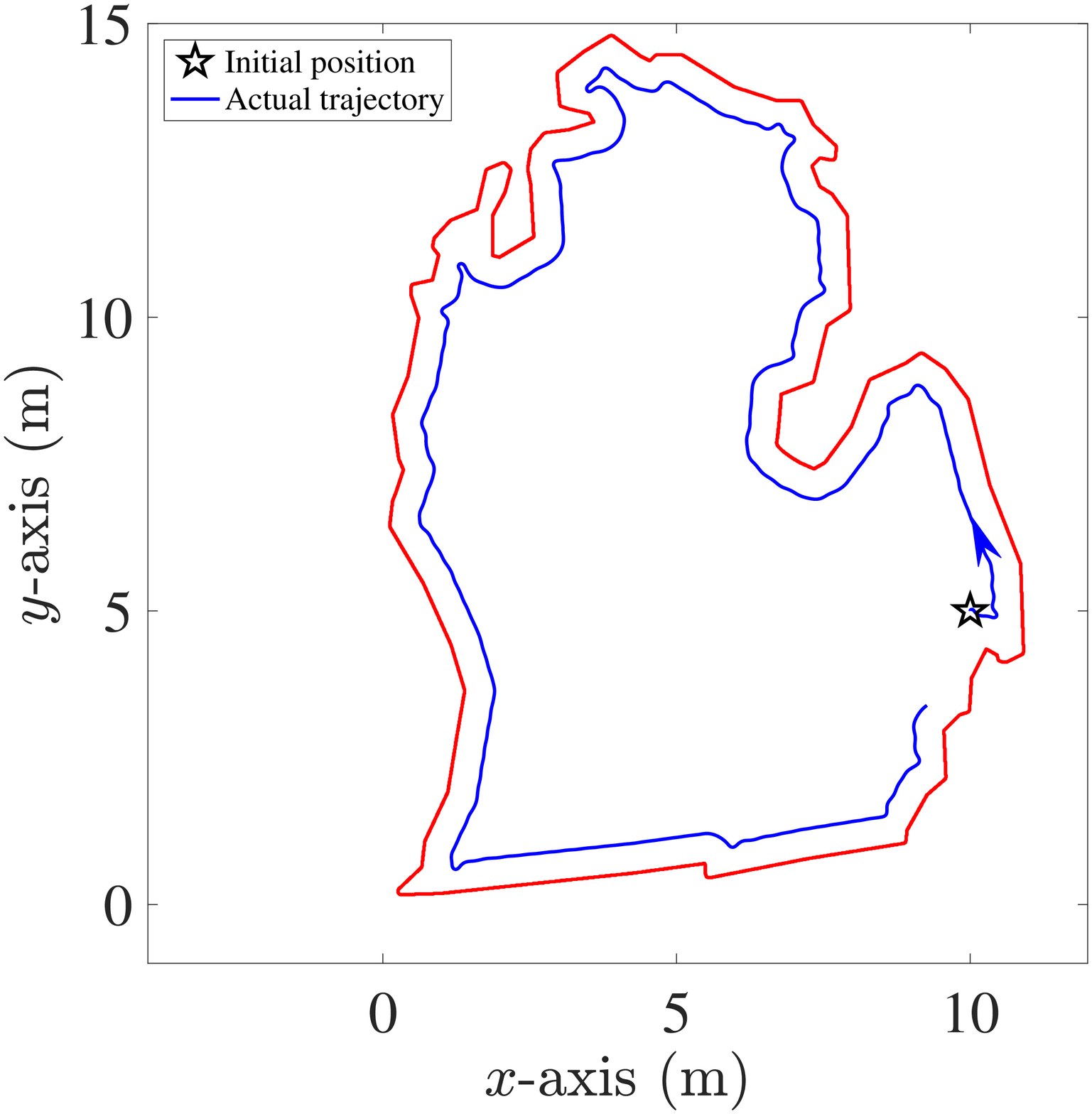}\\
{\bf(b)}\\
\includegraphics[width=0.98\columnwidth]{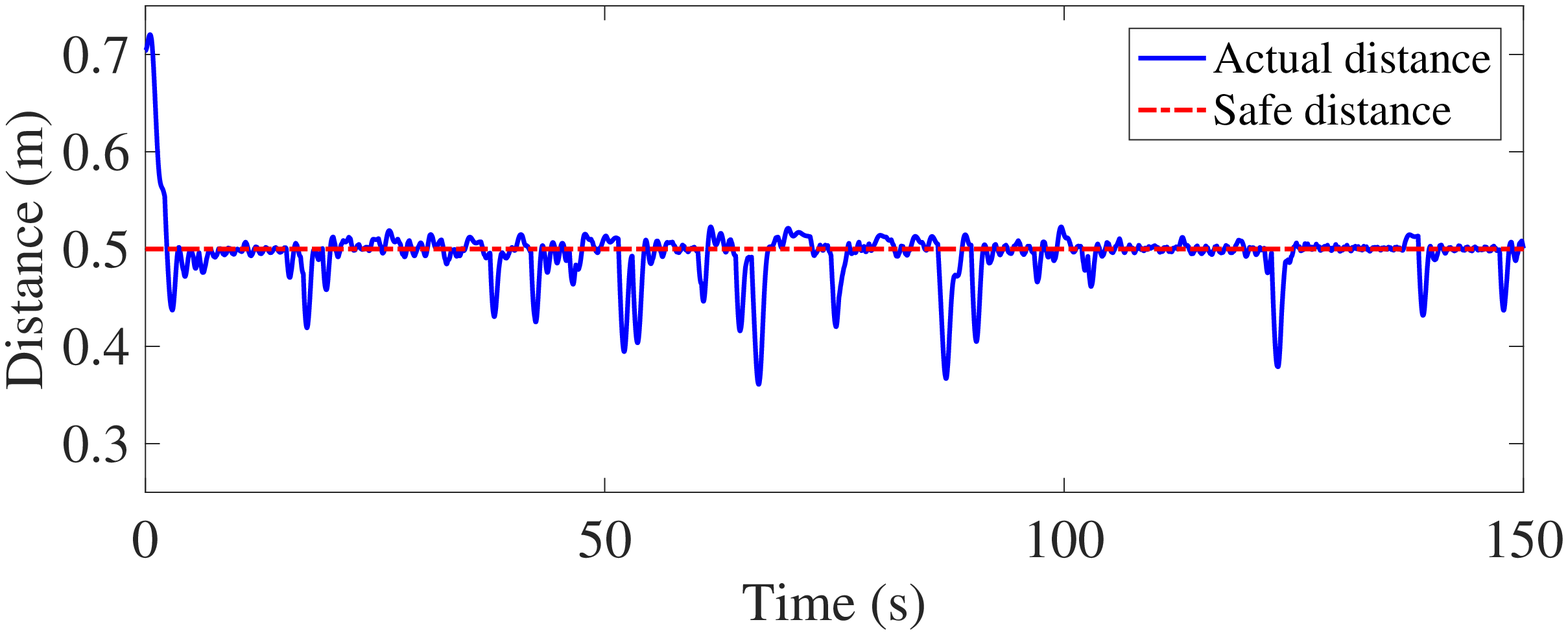}
\end{center}
\vspace{-3mm}
\caption{{\bf (a)} The robot performs border patrol with a constant linear velocity while maintaining a safe distance with the border shown in red. {\bf (b)} Robot's distance from the border. The average robot's distance from the border is $0.49$~m. In $93\%$ of the time, the robot's distance remains above $0.45$~m.}
\label{fig:BPpos}
\end{figure}

As shown in Fig.~\ref{fig:BPpos}(a), the safety algorithm used for geo-fencing can handle border patrol applications, where the linear speed is fixed at $0.5\pi$~m/s. As shown in Fig.~\ref{fig:BPpos}(b), the robot's distance is maintained very close to the safe distance, $\bar{d}_o=0.5$~m. In fact, in $93\%$ of the time, the safe distance remains above $0.45$~m. Tighter control bounds can be achieved by recalibrating the safety control gains, which may cause control over-actuation. In the case studies, the safety algorithm uses counterclockwise evasive maneuvers. Similar results can be produced for clockwise safety control.

\section{Conclusions}\label{sec:con}

The proposed variable structure control successfully isolates environmental properties from the performance criteria. It is shown that the derived nominal safety model adequately handles various safety-critical scenarios such as collision avoidance and geofencing with minimal information from the environment. The control structure comprises trajectory tracking, safety control, and a supervisory logic. Moreover, the super-twisting algorithm not only guarantees stability and safety but also facilitates control implementation. The proposed method dramatically simplifies the design steps and reduces the computation burden of the control compared to popular methods such as potential fields or path planning. Regardless of the number of obstacles or mobile robots, the algorithm's core components remain the same. Control bounds are obtained for the transformed systems, which facilitate adding features like integrator anti-windup to improve control performance. The results of four different case studies, conducted under realistic system constraints, verified the effectiveness of the proposed method. Future work extends the proposed method to unmanned aerial vehicles.


\bibliographystyle{IEEEtran}

\end{document}